# Enhancing Activity Recognition After Stroke: Generative Adversarial Networks for Kinematic Data Augmentation

**Aaron J Hadley [1] and Christopher L Pulliam [2,3]***

[1] Hadley Research, LLC, South Euclid, OH 44121, USA; aaronjhadley@gmail.com (A.J.H.)
[2] Department of Biomedical Engineering, Case Western Reserve University, Cleveland, OH 44106, USA
[3] Louis Stokes Cleveland Department of Veterans Affairs Medical Center, Cleveland, OH 44106, USA
* Correspondence: pulliam@case.edu

**Abstract:** The generalizability of machine learning (ML) models for wearable monitoring in stroke rehabilitation is often constrained by the limited scale and heterogeneity of available data. Data augmentation addresses this challenge by adding computationally derived data to real data to enrich the variability represented in the training set. Traditional augmentation methods, such as rotation, permutation, and time-warping, have shown some benefits in improving classifier performance, but often fail to produce realistic training examples. This study employs Conditional Generative Adversarial Networks (cGANs) to create synthetic kinematic data from a publicly available dataset, closely mimicking the experimentally measured reaching movements of stroke survivors. This approach not only captures the complex temporal dynamics and common movement patterns after stroke, but also significantly enhances the training dataset. By training deep learning models on both synthetic and experimental data, we enhanced task classification accuracy: models incorporating synthetic data attained an overall accuracy of 80.0%, significantly higher than the 66.1% seen in models trained solely with real data. These improvements allow for more precise task classification, offering clinicians the potential to monitor patient progress more accurately and tailor rehabilitation interventions more effectively.

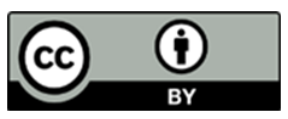

## 1. Introduction

Stroke is a leading cause of severe long-term disability among older adults in the United States. Approximately 9.4 million Americans live with residual neurological deficits caused by a stroke [1]. Despite recent advances in acute stroke management [2], approximately 60% of patients experience upper limb motor deficits, with only 5% to 20% demonstrating full recovery at six months post-stroke [3–6]. Many hemiparetic patients exhibit reduced elbow extension and shoulder flexion, which can impede reach, grasp, and transport movements [7,8]. Reduced upper limb capacity leads to dependence in daily activities and a lower quality of life for patients and caregivers, which, combined with an aging population, increases the urgency for innovative rehabilitation strategies [9,10].

Validated assessment scales, such as the Fugl-Meyer Motor Assessment (FMMA) and the Action Research Arm Test, evaluate upper limb motor impairment after stroke [11–14]. While these scales are responsive to recovery, they predominantly offer clinic-based insights, lacking data on patients' everyday performance. The assumption that clinical improvements will translate to better daily use overlooks learned non-use, where stroke survivors may not fully utilize their paretic limb in daily tasks despite clinical gains [15–17]. For instance, a patient able to grasp objects in a clinical setting might still avoid using their affected limb for daily tasks, a behavior that traditional assessments may not fully capture. The primary method to measure real-world upper-limb use has been through self-report questionnaires, such as the Motor Activity Log [18]. While these measures allow us to explore the relationship between capacity and performance, they do have inherent weaknesses as they rely on self-report, which can be confounded by issues with recall and perceptual difficulties in stroke survivors [19]. Ideally, new assessment methods would offer continuous, unobtrusive monitoring, providing a comprehensive view of the patient's real-world limb use and facilitating more dynamic adjustments to rehabilitation plans.

Body-worn motion sensors provide an opportunity for non-invasive, objective, and accurate observation of patients' movements during daily life [20–22]. The use of wearables in upper extremity stroke rehabilitation can be categorized into three main applications [23]: detecting functional or goal-directed movements [24–26], identifying motor impairment and functional limitations (e.g., predicting FM scores [27,28]), and quantifying real-world use [29–31]. Machine learning (ML) algorithms have been an important approach to advancing these applications towards clinical adoption but face a significant limitation in terms of performance on data from patients not in the training set (i.e., poor generalizability) [32,33]. This issue with generalizability has been attributed to heterogeneity of post-stroke movement patterns and the sparsity of available data. While developing patient-specific models can improve performance [32], the requirement to collect data on each user may pose a barrier to widespread clinical adoption. In contrast, data augmentation has emerged as a promising solution that artificially increases dataset sizes [34]. One approach is to perform statistical transformations of existing data, such as noise injection or time warping, to create new examples. Rotation, translation, and scaling of inertial measurement unit (IMU) signals have been employed to enhance activity classification accuracy in stroke patients on independent test data by 5-10% [35]. However, these methods often do not yield realistic training examples. A more sophisticated approach, demonstrated to be superior in medical imaging applications [36], involves synthesizing data that more accurately reflects real-world conditions using emerging generative artificial intelligence (AI) architectures [37,38].

Here, we evaluate the use of generative adversarial networks (GANs) to simulate post-stroke reaching movement patterns. While GANs have been used most extensively in imaging, recent evaluations include synthetic data augmentation in human motion analysis, including applications in motion capture [39,40] and Parkinson's disease [41,42]. Our work adds to this literature by evaluating: 1) the feasibility of using GANs to generate synthetic post-stroke upper extremity movement trajectories, and 2) the impact of augmenting ML training with synthetic data on activity recognition performance. The remainder of this paper is organized as follows. Section 2 discusses the source dataset used, the GAN architecture and training strategy, our approach to evaluating synthetic movements, and the classification problem we used to evaluate synthetic data augmentation. Section 3 presents the results of the simulations as well as the impact of augmented data on classification performance. Section 4 concludes with a discussion of the implications, limitations, and future directions.

## 2. Materials and Methods

### *2.1. Source Dataset*

Schwarz et al. used a wearable motion capture system (Xsens MVN Awinda, Xsens Technologies, Enschede, The Netherlands) to collect data on upper extremity kinematics from stroke survivors and healthy controls engaged in naturalistic daily living activities [43]. Details of the data collection protocol are described in detail elsewhere and summarized here. The study recorded joint angles and limb segment positions as participants performed a series of gesture and grasp movements. Gesture movements involved broad motor skills, such as arm lifting and reaching, that do not require object manipulation, while grasp movements involved more detailed hand and finger actions for handling objects. Each movement was executed up to three times with both the right and left limbs. A subset of these kinematic data is publicly available [44] and was used in our study. We analyzed trials from the most affected limb of stroke participants and the dominant limb in controls. Despite some limitations, such as the absence of trial-level annotations for movement quality and the lack of data from participants with severe impairment, these analyses allowed us to address basic feasibility questions regarding the use of GANs for data augmentation in recognizing activities performed by stroke survivors. **Table 1** details the demographic and clinical characteristics of the subjects. Subjects with an upper extremity FM score (FMMA-UE maximum score: 66) of less than 25 were considered "severe," between 25 and 53 were considered "moderate," and greater than 53 were considered "mild" impairment.

**Table 1**. Clinical and demographic summary.

| | **Controls** (n = 5) | **Mild Stroke** (n = 7) | **Moderate Stroke** (n = 12) | **Severe Stroke** (n = 1) |
|---|---|---|---|---|
| **Age** (years)[a] | 65 (52 – 68) | 55 (53 – 57) | 62 (58 – 70) | 74 |
| **Months Post Stroke**[a] | - | 21 (19 – 29) | 15 (8 – 54) | 12 |
| **FMMA-UE**[a] | - | 55 (54 – 60) | 42 (38 – 46) | 21 |
| **Sex** | m = 3 ∣ f = 2 | m = 6 ∣ f = 1 | m = 8 ∣ f = 4 | m = 1 ∣ f = 0 |
| **Trial Count** | 471 | 536 | 1149 | 90 |

[a] median (interquartile range); *FMMA-UE* Fugl-Meyer Motor Assessment of the Upper Extremity

*2.2. Data Selection and Pre-Processing*

A selection of ten tasks (T02: Distal Thumb Down, T03: Overhead, T04: Lateral, T06: Distal Palm Up, T08: Stop Gesture, T10: Hand to Mouth, T16: Grab and Bite Apple, T18: Move Cup, T19: Move Tray, T28: Move Tennis Ball) demonstrating various arm motions were chosen. This study focused on tasks that were similar in duration and involved a motion pattern of moving away from and then returning to a starting position. Other tasks, such as those involving grabbing an object and performing a more complex action (e.g., brushing teeth), were excluded due to their longer duration and additional movement complexity. For the design of the model, it was essential that all tasks were of comparable length. While minor variations were adjusted by padding or cropping, the longer tasks would have required either excessive cropping or significant padding, potentially introducing inconsistencies. As a result, we prioritized the shorter tasks with more uniform durations in this initial feasibility study.

Recordings from the control subjects (bilateral) and stroke subjects (more affected side) were selected. Given the imbalance in the source dataset, recordings from stroke survivors were categorized into two groups instead of three – "Mild" or "Moderate+Severe" – based on the recorded FMMA-UE score using a threshold of 42. This categorization resulted in 30 categories (i.e., classes): ten tasks across three distinct impairment levels, producing a total of 596 real reaching trials for model training and testing. Since the trials varied in length, the recordings were either cropped to 300 datapoints (5 seconds at a 60 Hz sample rate) or extended by repeating the final value to ensure identical length in subsequent analyses. Nine kinematic degrees of freedom (DOF) were selected to measure the ability of the GAN to simulate realistic movement trajectories: position (x, y, z) and orientation (z) of the T8 marker as measures of the trunk, as well as the shoulder (x, y, z) and elbow (x, y) joint angles. The recordings were normalized by converting T8 position changes into centimeters and shoulder/elbow motions into radians.

*2.3. GAN Model Development*

GANs are a class of ML frameworks where two neural networks, the generator and the discriminator, compete against each other. The generator creates synthetic data from random noise, while the discriminator evaluates whether the data is real or synthetic. In our approach, the generator network is trained on a dataset consisting of upper extremity kinematic data. The generator receives random noise as input and produces synthetic signals that mimic the kinematic data. The synthetic outputs and the original dataset are then input into the discriminator, which attempts to distinguish between real and synthetic data. The discriminator provides feedback to the generator as a loss function. This feedback is used to iteratively adjust the generator's parameters, enhancing its ability to produce realistic synthetic data. Concurrently, the discriminator improves its accuracy in distinguishing real data from synthetic data by learning from previous iterations. This adversarial training process results in the improvement of both networks.

Here, we implemented a Conditional GAN (cGAN), wherein class labels are provided as additional inputs to both the generator and discriminator [45]. This enables the generation of synthetic data conditioned on class information, improving the quality and relevance of the generated data for each category. This approach allows us to generate synthetic kinematic data for each of the 30 task/impairment categories defined in our dataset. The generated data is then used to augment the training set of subsequent ML activity recognition models.

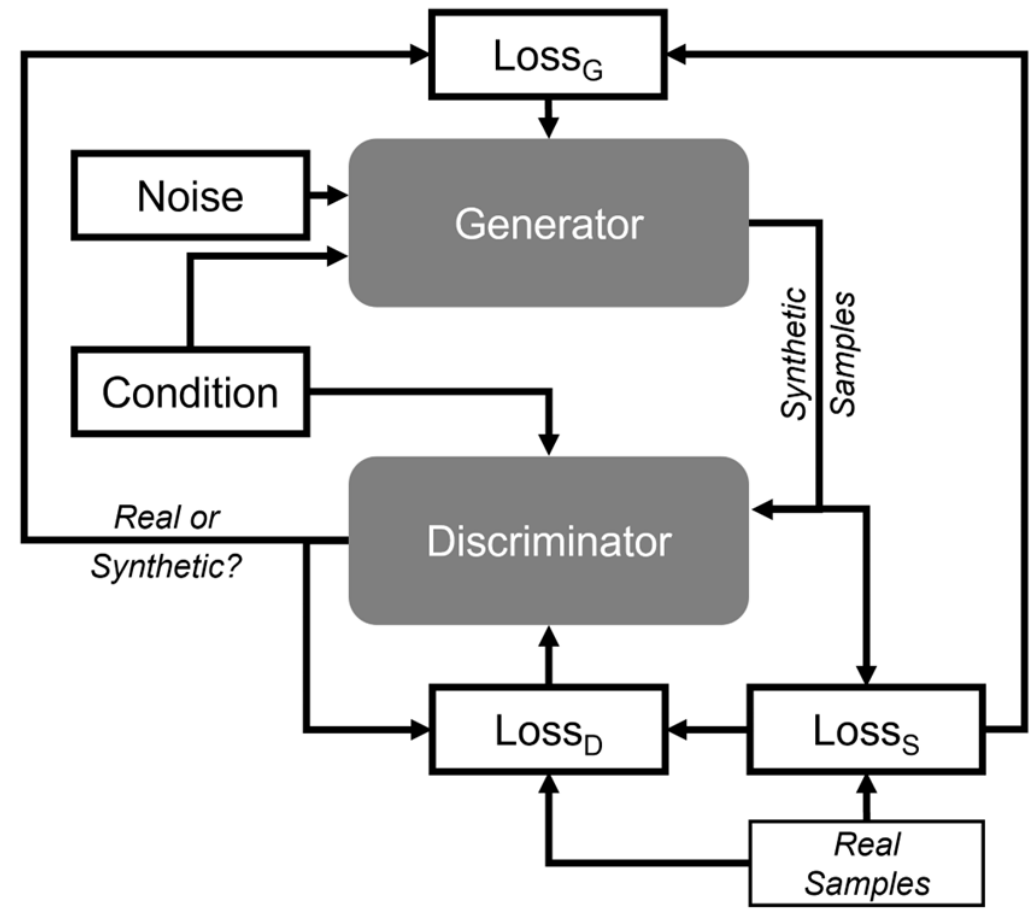

**Figure 1**. Diagram of the cGAN training framework.

In preliminary evaluations, the synthetic data generated with a standard cGAN exhibited elevated power in frequencies greater than 2 Hz. To address this, the standard cGAN architecture was modified to include a spectral loss function (**Figure 1**). This loss function, the squared difference of the Fast Fourier Transforms (FFTs) of real and synthetic samples, provided additional input to the generator to reduce differences in spectral energy between the two signals, making them more similar [42]. The generator inputs (i.e., noise vector and class label) are concatenated and passed through a dense layer followed by batch normalization, ReLU activation, and dropout with a rate of 0.4. The output is then reshaped for further processing through two upsampling blocks, each consisting of an upsampling layer followed by a Conv1D layer with 256 and 128 filters, respectively, both using a kernel size of 15, accompanied by batch normalization, ReLU activation, and dropout. The output size is adjusted with zero-padding or cropping to match the target sample length of 300, and a final Conv1D layer with 9 filters produces the required sensor channels. Additionally, a minibatch discriminator was implemented in training to ensure that the outputs from the generator were diverse, thereby addressing mode collapse [46]. Once the cGAN was trained using the Adam optimizer with a learning rate of 0.0003 and a beta_1 value of 0.5, a 2 Hz lowpass filter was applied to its output, and t-distributed Stochastic Neighbor Embedding (t-SNE) [47] was used for dimensionality reduction and qualitative visualization of how well the generated synthetic data approximated the real data. The range of motion for each DOF, a kinematic feature used to quantify stroke-related motor impairment [43], was employed to quantitatively compare the synthetic and real data. T-tests were conducted to compare the distributions of these ranges.

### *2.4. Data Augmentation*

To evaluate the impact of synthetic data augmentation, we added synthetic samples generated using cGANs to the training set for a task classification model. A fully convolutional network (FCN) illustrated in **Figure 2** was trained to classify the 10 tasks based on 5-second samples of the 9 kinematic signals. The first layer is a one-dimensional convolutional layer with 64 filters, a kernel size of 3, and ReLU activation, followed by a max-pooling layer with a pool size of 2. The second convolutional layer consists of 128 filters with a kernel size of 3 and ReLU activation, again followed by a max-pooling layer with a pool size of 2. The output is then flattened and passed through a fully connected layer with 100 units and ReLU activation. To prevent overfitting, a dropout layer with a rate of 0.5 is applied before the final dense layer, which contains units corresponding to the number of categories and uses softmax activation for classification. The model is compiled using the Adam optimizer with sparse categorical cross-entropy for loss calculation and accuracy as the evaluation metric.

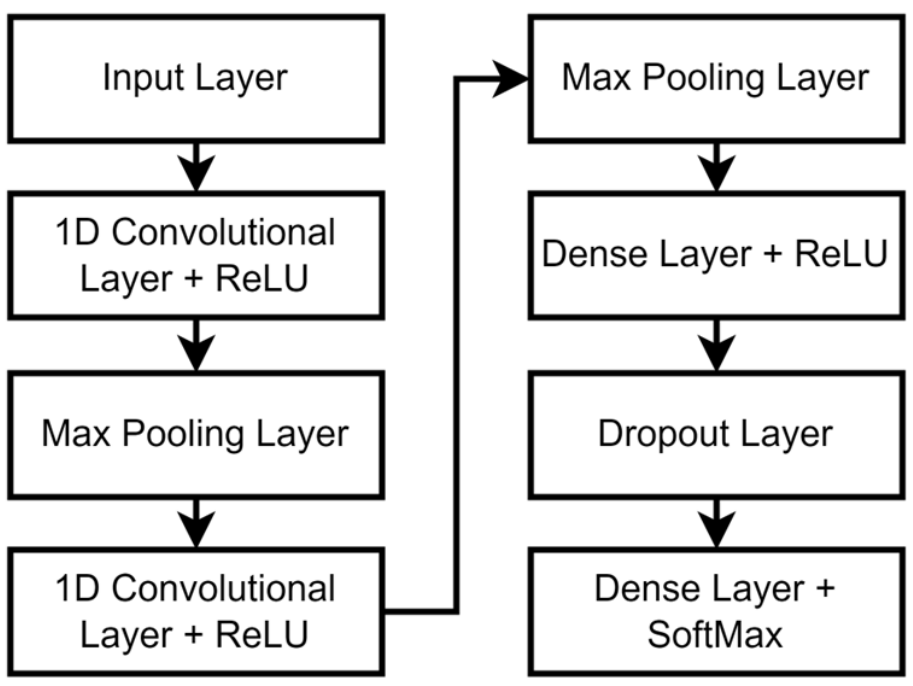

**Figure 2**. Diagram of the FCN architecture.

A five-fold cross-validation (CV) was conducted to evaluate model generalizability. This CV process was repeated a total of five times: once with the training set composed only of real data, and then with the training set in each fold augmented with synthetic data at varying ratios of real to synthetic data, specifically 2:1, 1:1, 2:3, and 1:2. Precision, recall, F1 score, and accuracy were used as metrics to quantify the effect of synthetic data augmentation on model performance.

## 3. Results

### *3.1. Synthetic Data Generation*

Figure 1 shows line plots comparing representative examples of real and synthetic movements for the “Grab and Bite Apple” tasks. The top two panels correspond to the x position of the T8 bony landmark, while the bottom two panels correspond to the forearm pronation-supination joint angle. Impairment level is shown by column. Within each panel, different colored lines illustrate different task instances within the respective categories. Qualitatively, synthetic samples shown in **Figure 3** follow the patterns observed in the real data. For example, a commonly used compensatory movement strategy is excessive movement of the trunk to accommodate for a diminished range of motion of the paretic arm [48]. This strategy is reflected in the real data, which show larger displacement of T8 for stroke survivors than for controls (row 2) and is mirrored in the synthetic data (row 1). The real data also reflect a reduced range of forearm pronation-supination motion as a function of impairment level (row 4) that is mirrored in the synthetic data (row 3).

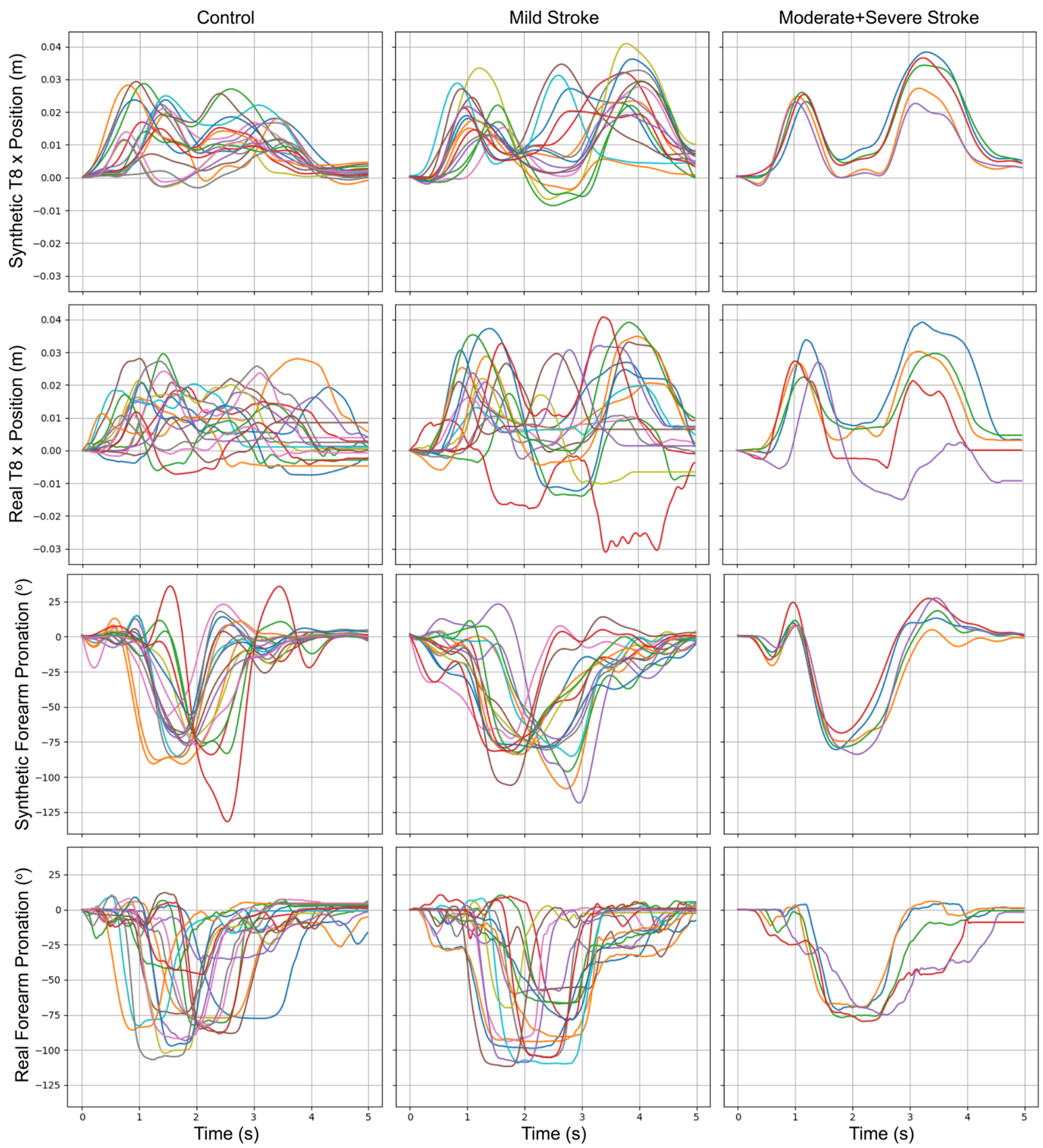

**Figure 3**. Representative synthetic and real movement trajectories during task 16 "grab and bite apple" for controls (left), mild stroke impairment (middle), and moderate+severe stroke impairment (right). The top two panels show the x-coordinate of the T8 bony landmark. The bottom two panels show the forearm pronation supination angle. Different colored lines in each panel represent different subjects.

A two-dimensional t-SNE plot of the entire dataset is included in **Figure 4**. Data points representing various tasks and impairment levels form distinct clusters, indicating that both real and synthetic data maintain task- and impairment-specific characteristics. P-values from t-tests comparing the range of motion for the real and synthetic samples are shown in **Table 2**. Tasks are listed by row and DOF by column. Each cell provides the result for control (top), mild (middle), and moderate+severe (bottom) impairment. While some similarities were observed in the ROM between real and synthetic trials across tasks, significant deviations were apparent, particularly in the moderate-to-severe stroke group. These deviations were most pronounced in the shoulder and elbow.

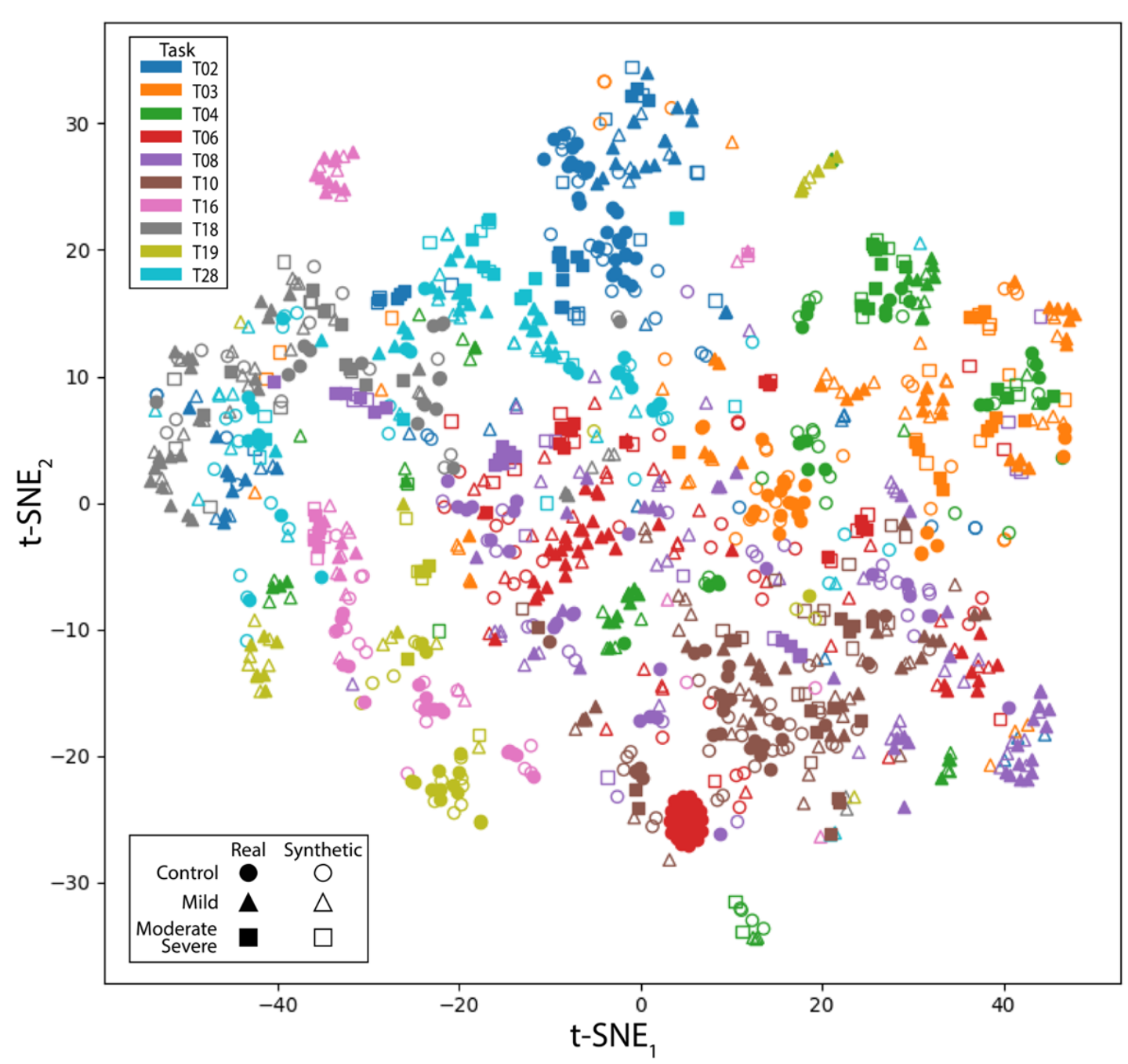

**Figure 4**. t-SNE visualization of the distributions of real and synthetic data. Tasks are indicated by color. Marker shape indicates impairment level, with circles representing control subjects, triangles representing stroke subjects with mild impairment, and squares representing stroke subjects with moderate or severe impairment. Open markers represent synthetic samples, while filled markers represent real samples.

**Table 2.** Significance levels for range of motion comparisons by task and degree of freedom.

| | *T8x* | *T8y* | *T8z* | $T8\theta$ | $S\theta x$ | $S\theta y$ | $S\theta z$ | $E\theta x$ | $E\theta y$ |
|---|---|---|---|---|---|---|---|---|---|
| | 0.201 | 0.129 | 0.203 | **0.009** | 0.055 | 0.141 | 0.361 | 0.151 | 0.648 |
| *T02* | 0.390 | 0.255 | 0.059 | 0.307 | **0.007** | 0.068 | 0.134 | **0.000** | 0.478 |
| | 0.476 | **0.015** | 0.122 | 0.735 | **0.014** | 0.220 | 0.927 | **0.000** | **0.013** |
| | 0.366 | 0.062 | 0.575 | 0.281 | 0.346 | 0.092 | 0.336 | 0.624 | 0.885 |
| *T03* | 0.773 | 0.681 | **0.030** | 0.515 | **0.019** | **0.022** | **0.013** | 0.305 | **0.002** |
| | 0.678 | 0.095 | 0.361 | **0.021** | **0.040** | 0.097 | **0.002** | 0.235 | **0.029** |
| | 0.949 | 0.319 | 0.214 | 0.292 | **0.012** | 0.423 | **0.029** | **0.000** | 0.166 |
| *T04* | 0.920 | **0.016** | 0.182 | 0.068 | **0.000** | 0.821 | 0.193 | **0.021** | 0.697 |
| | **0.001** | 0.884 | 0.549 | 0.302 | 0.267 | 0.721 | 0.456 | **0.000** | 0.466 |
| | **0.003** | **0.000** | **0.000** | **0.000** | **0.015** | **0.015** | **0.000** | **0.016** | **0.001** |
| *T06* | 0.131 | 0.336 | 0.223 | 0.978 | 0.998 | 0.929 | 0.272 | 0.112 | 0.874 |
| | 0.095 | 0.155 | 0.623 | **0.002** | **0.000** | **0.019** | **0.000** | **0.043** | **0.431** |
| | 0.319 | 0.329 | 0.491 | 0.423 | 0.202 | 0.701 | 0.152 | 0.432 | 0.161 |
| *T08* | 0.364 | 0.842 | 0.805 | 0.203 | 0.839 | **0.014** | **0.047** | 0.662 | 0.187 |
| | 0.341 | 0.100 | 0.101 | 0.488 | **0.005** | **0.000** | 0.943 | 0.145 | **0.030** |
| | 0.678 | 0.330 | 0.235 | 0.883 | **0.001** | 0.524 | 0.767 | 0.645 | **0.012** |
| *T10* | **0.002** | 0.632 | 0.583 | 0.365 | 0.584 | 0.116 | 0.320 | **0.002** | **0.020** |
| | 0.993 | **0.022** | 0.306 | 0.110 | **0.003** | **0.000** | **0.006** | **0.010** | **0.043** |
| | 0.816 | 0.232 | 0.532 | **0.021** | **0.000** | **0.011** | **0.047** | **0.005** | 0.336 |
| *T16* | **0.043** | 0.654 | 0.976 | 0.670 | **0.000** | **0.001** | **0.003** | **0.018** | 0.549 |
| | 0.900 | 0.459 | **0.000** | 0.710 | **0.000** | **0.000** | **0.000** | **0.000** | **0.000** |
| | 0.139 | 0.067 | 0.485 | 0.372 | 0.198 | **0.002** | **0.001** | **0.000** | **0.005** |
| *T18* | **0.001** | 0.051 | 0.337 | 0.289 | **0.046** | **0.000** | **0.001** | **0.000** | **0.009** |
| | 0.260 | 0.240 | **0.047** | 0.450 | **0.000** | 0.063 | **0.000** | **0.000** | **0.000** |
| | 0.691 | 0.822 | 0.102 | 0.202 | **0.035** | **0.002** | 0.052 | **0.000** | **0.000** |
| *T19* | **0.014** | 0.056 | 0.784 | **0.005** | 0.117 | 0.417 | **0.025** | 0.457 | **0.000** |
| | **0.008** | **0.000** | **0.001** | **0.000** | **0.000** | **0.000** | **0.000** | 0.510 | **0.000** |
| | 0.499 | 0.777 | 0.402 | 0.521 | **0.000** | **0.024** | **0.001** | **0.000** | 0.420 |
| *T28* | 0.147 | 0.917 | 0.627 | 0.059 | **0.000** | 0.864 | 0.209 | 0.193 | **0.041** |
| | 0.532 | 0.294 | 0.166 | 0.905 | **0.000** | **0.000** | **0.002** | **0.000** | 0.886 |

### 3.2. *Impact of Synthetic Data Augmentation*

To evaluate the impact of synthetic data augmentation, we augmented the training set for the FCN model with synthetic samples generated using cGANs. **Figure 5** presents two confusion matrices comparing classification results: one for the model trained exclusively on real data (top) and the other for the model trained with a 1:1 ratio of real and synthetic samples (bottom). Each cell represents the percentage of real data points correctly or incorrectly classified across the 10 selected tasks. The matrix for the augmented training set shows reduced misclassifications.

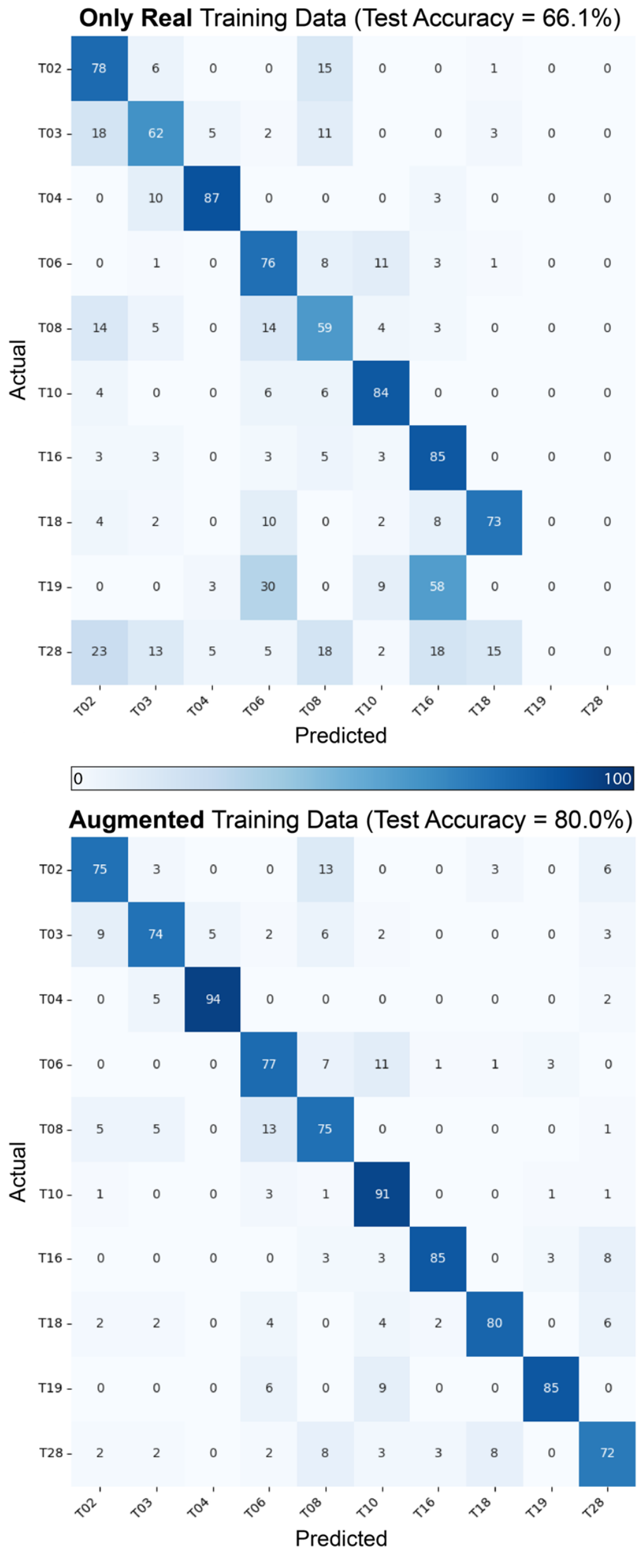

**Figure 5.** Confusion matrices illustrating the impact of synthetic data augmentation on classification of task and impairment level. The top matrix results from training with real data alone, while the bottom matrix results from training with real and synthetic data in a 1:1 ratio. In both cases, the models were tested with real data.

**Table 3** shows the precision, recall, and F1 values for the five compositions of training data. The performance metrics varied across different ratios of real to synthetic data. Training on real data alone yielded the lowest performance across all metrics. As the proportion of synthetic data increased to a 2:1 and 1:1 ratio of real to synthetic data, there was a notable improvement in all metrics. However, performance declined when the ratio shifted to 2:3 and 1:2, with lower values observed across precision, recall, F1 score, and accuracy.

**Table 3**. Classifier performance metrics.

| | *Precision* | *Recall* | *F1 Score* | *Accuracy* |
|---|---|---|---|---|
| *Real Only* | 0.575 ± 0.036 | 0.661 ± 0.042 | 0.608 ± 0.039 | 0.661 ± 0.042 |
| *2:1* | 0.707 ± 0.054 | 0.688 ± 0.060 | 0.686 ± 0.061 | 0.688 ± 0.060 |
| *1:1* | 0.809 ± 0.022 | 0.800 ± 0.024 | 0.801 ± 0.022 | 0.800 ± 0.024 |
| *2:3* | 0.618 ± 0.032 | 0.617 ± 0.026 | 0.604 ± 0.031 | 0.617 ± 0.026 |
| *1:2* | 0.629 ± 0.048 | 0.617 ± 0.043 | 0.604 ± 0.042 | 0.617 ± 0.043 |

## 4. Discussion

This study demonstrates that integrating synthetic kinematic data generated by cGANs into training can enhance the performance of ML models for activity monitoring in stroke rehabilitation. The synthetic data not only replicate the dynamics of post-stroke patient movements accurately but also enrich the training dataset, potentially reducing the reliance on extensive experimental data collection. This enriched dataset leads to notable improvements in precision, recall, F1 score, and overall accuracy, which are essential for accurately classifying tasks and impairment levels in stroke rehabilitation. Such enhancements are crucial for developing personalized rehabilitation strategies that are tailored to the recovery profiles of individual patients.

Qualitative analysis using line graphs to compare synthetic and real trajectories (**Figure 3**) shows that synthetic data capture general movement trends across various impairment groups while introducing additional variability. This variability could benefit the training process, potentially enhancing model robustness by equipping systems to handle real-world variations more effectively. Furthermore, t-SNE analysis (**Figure 4**) demonstrates that synthetic data effectively cluster tasks and impairment levels, indicating that these data preserve task-specific and impairment-specific characteristics and closely mirror the distribution and features of the real data. Confusion matrices (**Figure 5**) show a significant improvement in model accuracy with the inclusion of synthetic data in the training set. The test accuracy increased from 66.1% to 80.0%, and there were corresponding improvements in precision, recall, and F1 scores (**Table 3**). The ratio of real to synthetic data appeared to have an optimum at 1:1, as higher or lower ratios resulted in diminished performance, highlighting the importance of maintaining a balanced data augmentation strategy. The observed improvement in accuracy with augmented data is likely due to the additional variability introduced by synthetic samples, which helps the model generalize better to unseen data. In contrast, models trained solely on real data are more prone to overfitting, particularly when the dataset is small. The synthetic data provides a more diverse representation of possible movement patterns, improving the model's robustness in classifying tasks and impairment levels. Together, these results highlight the capability of data augmentation to mitigate the challenges posed by small and imbalanced datasets.

However, the focus on kinematic trajectories rather than IMU signals – more commonly used in stroke rehabilitation – marks a limitation of this study. Future efforts will explore adapting these methods to synthesize IMU signals and assess their effectiveness in model training. Additionally, the realism of the synthetic data was primarily assessed through qualitative means. Quantitative evaluation was limited to range of motion, and notable discrepancies were observed between real and synthetic data, especially in more severely impaired groups (**Table 2**). Parameters such as spectral arc length, number of velocity peaks, trunk displacement, range of motion, and inter-joint coupling interval have been shown to be useful for assessing the spatiotemporal aspects of upper limb movement behavior after stroke [43,49,50]. These parameters could provide a robust basis for quantitative comparisons of real and synthetic data, improving the realism of synthetic datasets. While the implementation of a 2 Hz lowpass filter helped reduce noise in the synthetic data, task-specific frequency tuning could further enhance the accuracy of the generated movement trajectories. Future work may explore adaptive filtering techniques to optimize synthetic data quality across varying movement tasks and improve overall model performance. Directly integrating these measures into customized loss functions might also improve the realism of the synthetic data, enhancing the model's ability to generalize to new, unseen data.

Another limitation arises from the source dataset's composition, which included only a single subject with severe motor impairment, requiring the combination of moderate and severe categories. This could limit the generalizability of findings across a full spectrum of stroke severity. Collecting a more balanced dataset in future research will enhance the applicability and accuracy of the findings. Another key limitation of this study was the lack of detailed FMMA-UE subscores, which could have provided insight into specific joint impairments. The use of only total FMMA-UE scores limited our ability to discern how deficits in particular joints or movement patterns contributed to overall impairment. Furthermore, the dataset did not include annotations on the quality of individual movements, such as compensatory strategies or abnormal kinematics that are often observed in stroke survivors. The absence of these annotations made it challenging to evaluate how well the synthetic data replicated these movement abnormalities. More granular annotations of movement quality could improve the accuracy and clinical relevance of future analyses, especially when assessing the realism of synthetic data. Moving forward, the inclusion of detailed movement quality metrics, such as joint-specific assessments or compensatory movement indicators, would enable more precise evaluations of both real and synthetic movement data, ultimately enhancing the clinical applicability of the models. Additionally, the source dataset did not include raw IMU data, which prevented us from assessing the feasibility of directly simulating these signals, given their importance in potential clinical applications. Generative modeling of IMU data presents additional challenges compared to joint-level kinematic data due to the higher dimensionality and number of signals produced by each sensor. IMUs capture multiple degrees of freedom, including acceleration and angular velocity across three axes, leading to a more complex dataset. Additionally, IMU data reflects faster temporal dynamics, which require fine-grained synthesis to capture rapid fluctuations in movement patterns. These factors complicate the generation of realistic synthetic IMU data, especially for movements involving compensatory strategies or subtle shifts in body position. Continued efforts will focus on validating synthetic IMU signals and examining how different ML architectures respond to data augmentation. A systematic evaluation of various models will help identify optimal training configurations to prevent overfitting. Comparing the efficacy of generative data augmentation with traditional statistical techniques [51] will further delineate the benefits and preferable contexts for each method [35].

## 5. Conclusions

This study demonstrates that using Conditional GANs to generate synthetic kinematic data significantly enhances task classification accuracy in stroke rehabilitation. By addressing data scarcity and improving generalizability, this approach offers a promising solution for more accurate monitoring and personalized rehabilitation interventions. This increased generalizability allows clinicians to apply these models more broadly, potentially facilitating personalized rehabilitation interventions without the burden of collecting large datasets for each individual, making the models more scalable and adaptable for clinical use.

**Author Contributions:** Conceptualization, C.L.P.; methodology, A.J.H.; software, A.J.H.; validation, A.J.H.; formal analysis, A.J.H.; investigation, A.J.H. and C.L.P.; resources, C.L.P.; data curation, A.J.H. and C.L.P.; writing—original draft preparation, C.L.P.; writing—review and editing, A.J.H.

and C.L.P.; visualization, A.J.H. and C.L.P.; supervision, C.L.P.; project administration, A.J.H. and C.L.P.; funding acquisition, C.L.P. All authors have read and agreed to the published version of the manuscript.

**Funding:** This research was supported by internal funding from Case Western Reserve University.

**Data Availability Statement:** This study involved secondary analyses of data that are openly available in Zenodo at 10.5281/zenodo.3713449 [44].

**Conflicts of Interest:** The authors declare no conflicts of interest